\documentclass{article}
\usepackage{ijcai25}

\usepackage{times}
\usepackage{soul}
\usepackage{url}
\usepackage[hidelinks]{hyperref}
\usepackage[utf8]{inputenc}
\usepackage[small]{caption}
\usepackage{graphicx}
\usepackage{amsmath}
\usepackage{amsthm}
\usepackage{booktabs}
\usepackage{algorithm}
\usepackage{algorithmic}
\usepackage[switch]{lineno}

\usepackage{amssymb}
\usepackage{graphicx}
\usepackage{adjustbox}
\usepackage{amsmath}
\usepackage{balance}
\usepackage{soul} 
\usepackage{color, xcolor} 
\usepackage{makecell}
\usepackage{multirow}
\usepackage{enumitem}
\usepackage{enumerate}
\usepackage{colortbl}
\usepackage{xspace}
\definecolor{1}{RGB}{255,146,146}
\definecolor{2}{RGB}{255,204,153}
\definecolor{3}{RGB}{255,255,153}
\definecolor{check}{RGB}{0,153,76}
\definecolor{fork}{RGB}{233,75,75}
\usepackage{array}
\newcommand{\PreserveBackslash}[1]{\let\temp=\\#1\let\\=\temp}
\newcolumntype{C}[1]{>{\PreserveBackslash\centering}p{#1}}
\newcolumntype{R}[1]{>{\PreserveBackslash\raggedleft}p{#1}}
\newcolumntype{L}[1]{>{\PreserveBackslash\raggedright}p{#1}}
\usepackage{svg}

\title{PEP-GS: Perceptually-Enhanced Precise Structured 3D Gaussians for View-Adaptive Rendering}

\author{
Junxi Jin$^{1}$
\and
Xiulai Li$^{1}$\footnote{Corresponding Author}\and
Haiping Huang$^{1}$\and
Lianjun Liu$^{1}$\and \\
Yujie Sun$^{1}$\And
Logan LIU$^2$
\\
\affiliations
$^1$Hainan University, Haikou, China\\
$^2$The Hong Kong University of Science and Technology, Hong Kong SAR\\
\emails
\{junxijin, lixiulai01, huanghp, lianjun\_320, sunyujie\}@hainanu.edu.cn \\
bliubd@connect.ust.hk
}

\begin{document}

\maketitle

\begin{abstract}
 Recently, 3D Gaussian Splatting (3D-GS) has achieved significant success in real-time, high-quality 3D scene rendering. However, it faces several challenges, including Gaussian redundancy, limited ability to capture view-dependent effects, and difficulties in handling complex lighting and specular reflections. Additionally, methods that use spherical harmonics for color representation often struggle to effectively capture anisotropic components, especially when modeling view-dependent colors under complex lighting conditions, leading to insufficient contrast and unnatural color saturation. To address these limitations, we introduce PEP-GS, a perceptually-enhanced framework that dynamically predicts Gaussian attributes, including opacity, color, and covariance. We replace traditional spherical harmonics with a Hierarchical Granular-Structural Attention mechanism, which enables more accurate modeling of complex view-dependent color effects. By employing a stable and interpretable framework for opacity and covariance estimation, PEP-GS avoids the removal of essential Gaussians prematurely, ensuring a more accurate scene representation. Furthermore, perceptual optimization is applied to the final rendered images, enhancing perceptual consistency across different views and ensuring high-quality renderings with improved texture fidelity and fine-scale detail preservation. Experimental results demonstrate that PEP-GS outperforms state-of-the-art methods, particularly in challenging scenarios involving view-dependent effects and fine-scale details.

\end{abstract}

\vspace{-1.em}
\begin{figure}
    \centering
    \includegraphics[width=0.99\linewidth]{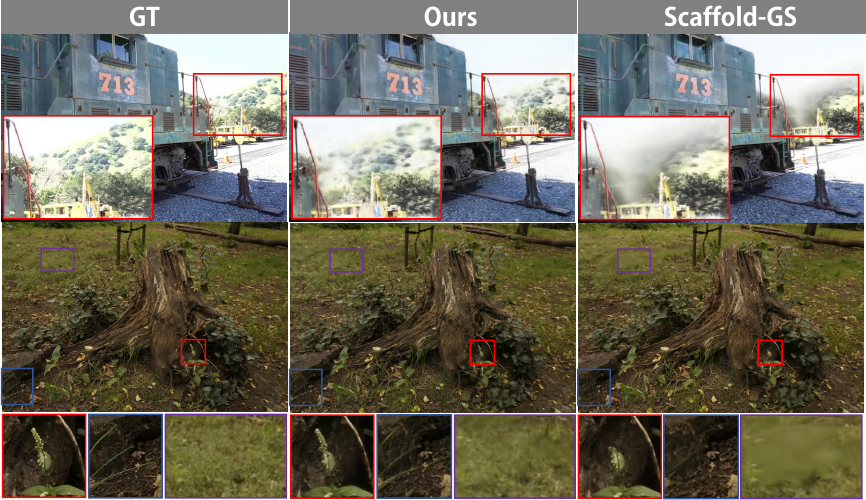}
    \caption{Comparison of PEP-GS against SOTA methods. Our method achieves better perceptual consistency and more accurate view-dependent effects across different viewing angles, particularly in scenes with complex lighting and textures}
    \label{fig:Intro}
\vspace{-1.5em}
\end{figure}

\section{Introduction}
\label{sec:intro}

The rendering of 3D scenes has long been a cornerstone of both computer vision and graphics research, with numerous applications across industries such as virtual reality~\cite{jiang2024vr,xie2024physgaussian,xu2023vr}, autonomous driving~\cite{Zhou_2024_CVPR,10.1145/3641519.3657448}, and drivable human avatars~\cite{qian20243dgs,qian2024gaussianavatars,zheng2024gps}. Traditional methods, such as meshes and point clouds~\cite{botsch2005high,munkberg2022extracting,yifan2019differentiable}, are efficient but often compromise visual quality, leading to discontinuities. In contrast, volumetric representations like Neural Radiance Fields (NeRF)~\cite{barron2022mipnerf360,mildenhall2021nerf,Barron_2023_ICCV} deliver photorealistic results but are computationally intensive, particularly for real-time rendering. Recently, 3D Gaussian Splatting (3D-GS)~\cite{kerbl3Dgaussians} has emerged as a promising solution, enabling high-quality, real-time rendering of scenes using 3D Gaussian primitives, thus bypassing the need for complex ray-sampling techniques.

Although 3D-GS demonstrates impressive rendering performance, it struggles with capturing complex view-dependent effects, especially in dynamic lighting environments. This limitation stems from the inherent constraints of low-order spherical harmonics (SH) used for color representation in 3D-GS, which are well-suited for modeling subtle, view-dependent phenomena but struggle to accurately represent anisotropic components. Due to their low-frequency nature, SH functions struggle to capture high-frequency, view-dependent details under dynamic lighting, which limits their effectiveness in complex scenes. Moreover, structured approaches like Scaffold-GS reduce Gaussian redundancy through hierarchical and region-aware scene representation but struggle with fine-scale details and texture modeling, leading to perceptual inconsistencies under complex lighting and reflections.

To address these challenges, we propose PEP-GS, a perceptually-enhanced framework for structured 3D Gaussian splatting inspired by previous work~\cite{lu2024scaffold,laparra2016perceptual,xu2024ela}. Similarly, we construct a sparse grid of anchor points, initialized from SfM points. For each neural Gaussian tethered to these anchors, PEP-GS dynamically predicts Gaussian attributes including opacity for transparency control, color for view-dependent appearance modeling, and covariance for spatial representation using a stable and precise framework. Specifically, we replace the traditional SH-based color representation with a Hierarchical Granular-Structural Attention mechanism, enabling more accurate modeling of view-dependent color effects. Additionally, we integrate Kolmogorov-Arnold Networks (KAN) for opacity and covariance estimation, enhancing the stability of the predictions. Building on this, we apply perceptual optimization techniques to improve the consistency of the rendered results. This approach refines the image details and texture fidelity, optimizing visual consistency across different views and significantly improving perceptual fidelity and fine-scale detail preservation in the final rendered images.

Through extensive experimentation, we demonstrate that our approach significantly outperforms existing state-of-the-art methods. As shown in Figure.~\ref{fig:Intro}, it shows particular strength in handling challenging scenarios such as view-dependent lighting effects, and intricate geometric details. Notably, our method maintains stability and consistently improves visual quality and perceptual consistency across novel views.

The primary contributions of this work are:
\begin{itemize}
    \item A perceptually-enhanced framework for structured 3D Gaussian splatting that significantly improves rendering quality in view-dependent scenarios while avoiding the premature removal of essential neural Gaussians, ensuring efficient scene representation.
    
    \item Novel neural architectures that effectively and stably predict key Gaussian attributes, addressing critical limitations in current Gaussian-based rendering approaches, particularly in terms of accuracy and stability under complex lighting conditions.

    \item Comprehensive experimental validation demonstrating substantial improvements in rendering quality across multiple datasets, particularly in challenging scenarios involving complex lighting and fine geometric details.
\end{itemize}

\section{Related work}
\label{sec:Related_work}
\subsection{Neural Scene Representations and 3D Gaussians}
Neural scene representation has evolved significantly from early MLP-based approaches to current state-of-the-art methods. NeRF~\cite{mildenhall2021nerf} pioneered the use of volumetric representations through coordinate-based neural networks, achieving high-quality results but suffering from slow rendering speeds. Subsequent works explored various acceleration techniques~\cite{cao2023real,chen2022tensorf,chen2023mobilenerf,reiser2021kilonerf,reiser2023merf,sitzmann2021light}. Although NeRF-based methods have made significant strides in volumetric rendering, achieving an optimal balance between visual fidelity and processing speed remains a major challenge. Meanwhile, point-based rasterization~\cite{franke2023vet,franke2024trips,kopanas2021point,xu2022point} approaches have attracted attention for their ability to deliver competitive rendering performance with improved computational efficiency. A breakthrough came with 3D Gaussian Splatting (3D-GS)~\cite{kerbl3Dgaussians}, which achieves real-time performance by explicitly modeling the scene with optimized 3D Gaussians and a differentiable rasterization pipeline. However, this can result in redundant Gaussians that attempt to fit every training view, negatively impacting robustness under significant view changes and challenging lighting conditions. The Scaffold-GS~\cite{lu2024scaffold} subsequently introduced a structured hierarchy of 3D Gaussians anchored by sparse SfM~\cite{sfm} points, effectively reducing redundant Gaussian usage and enhancing the representation's adaptability to varying levels-of-detail.

\subsection{View-dependent Appearance Modeling}
Accurately capturing high-frequency view-dependent phenomena (e.g., specular reflections) remains a key challenge in neural rendering. Existing volumetric methods, including NeRF derivatives, often rely on explicit or implicit encodings to handle view dependence~\cite{xu2022point,barron2022mipnerf360,mueller2022instant} but can struggle with glossy materials and realistic reflectance. In the 3D-GS family, spherical harmonics (SH) have been widely adopted for their computational efficiency~\cite{kerbl3Dgaussians,fridovich2022plenoxels}, yet they primarily excel at modeling subtle view-dependent effects and can introduce artifacts for complex lighting. Several approaches have attempted to overcome these limitations through various encoding schemes~\cite{verbin2022ref,ma2024specnerf,yang2024spec}; however, achieving perceptual consistency across different viewing angles while preserving geometric details remains challenging. Our work addresses these limitations through a perceptually-enhanced framework that better handles view-dependent effects and geometric details while maintaining rendering consistency across different views.

\section{Methodology} 
\label{sec:methods}
\begin{figure*}[h]
    \centering
    \includegraphics[width=0.95\textwidth]{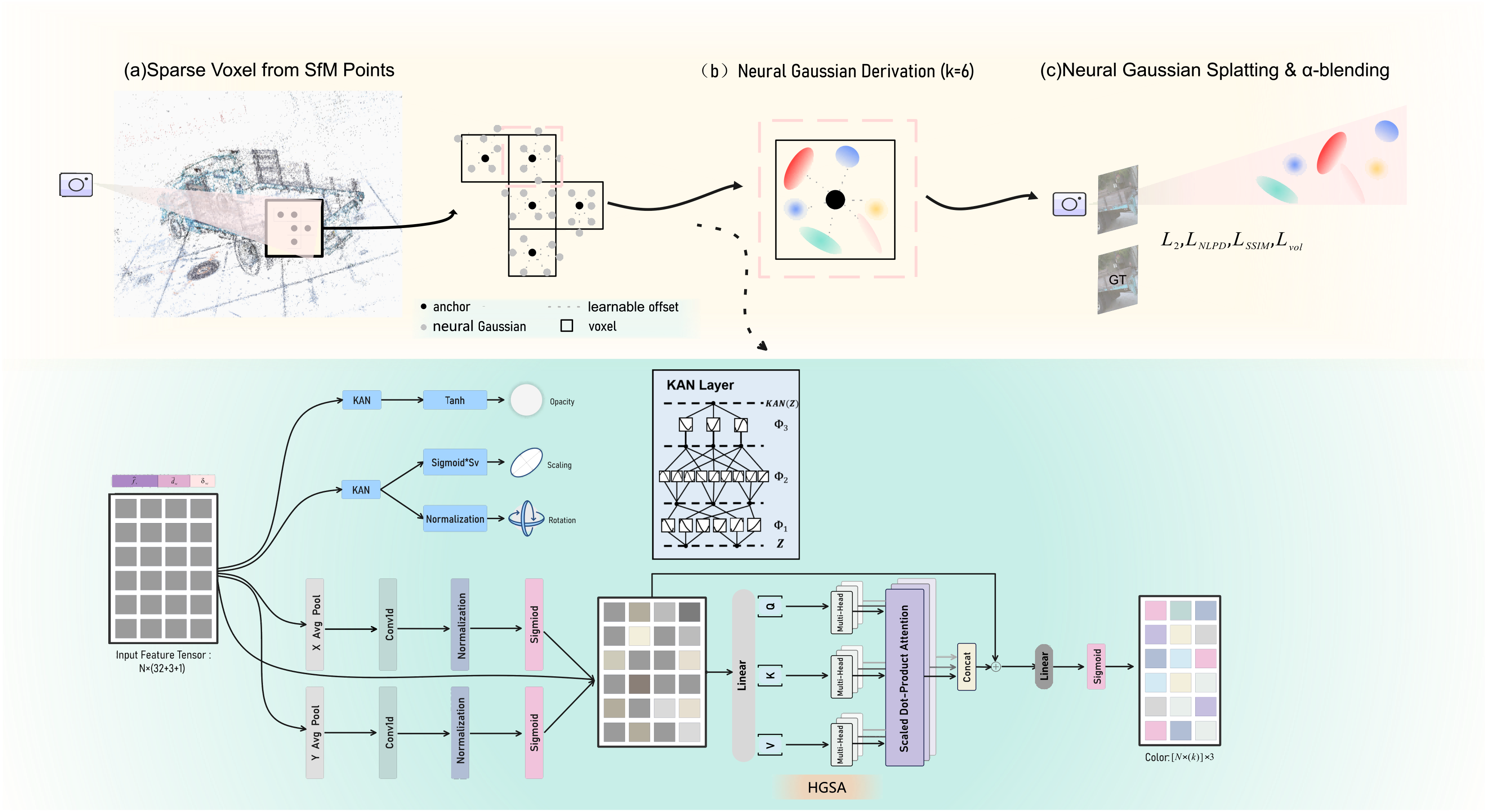}
    \vspace{-5pt}
    \caption{Overview of PEP-GS}
    \vspace{-15pt}
    \label{fig:overview}
\end{figure*}

The overview of our method is illustrated in Figure.~\ref{fig:overview} Starting with anchor points derived from SfM data, each anchor encodes localized scene information to guide tethered neural Gaussians, which dynamically predict attributes such as opacity, color, rotation, and scale based on the viewing angle. To improve the modeling of view-dependent lighting effects, we replace spherical harmonics with Hierarchical Granular-Structural Attention Mechanism to predict the color attribute, which enables more effective capture of local geometric features and complex view-dependent color changes. For opacity, rotation, and scale, we employ Kolmogorov-Arnold Networks~\cite{liu2024kan}, offering a more interpretable and precise framework for neural Gaussian attribute prediction. Finally, we incorporate the Normalized Laplacian Pyramid Distance (\(L_\text{NLPD}\)) as a perceptual loss function to ensure cross-view perceptual consistency, resulting in coherent, high-quality renderings.

\subsection{Preliminaries}
3D Gaussian Splatting~\cite{kerbl3Dgaussians} (3D-GS) models scenes using a collection of anisotropic 3D Gaussians, each defined by center position $\mu$, opacity $\alpha$, covariance matrix $\Sigma$, and spherical harmonic (SH) coefficients. Specifically, each 3D Gaussian is centered at $\mu$ with a covariance $\Sigma$:
\begin{equation}
    G(x) = e^{-\frac{1}{2}(x - \mu_i)^T \Sigma_i^{-1} (x - \mu_i)},
\end{equation}
where $x$ is an arbitrary position within the 3D scene. The positive semi-definite covariance matrix $\Sigma$ is constructed from scaling $S$ and rotation $R$ matrices as $\Sigma = RSS^T R^T$. Then 3D Gaussians $G$ are projected to 2D Gaussians $G'$ for rendering. Using depth-sorted 2D Gaussians, the tile-based rasterizer applies $\alpha$-blending to calculate pixel color $C(x')$:
\begin{equation}
    C(x') = \sum_{i=1}^{N} c_i \alpha_iG_i(x') \prod_{j=1}^{i-1} (1 - \alpha_j G'_j(x')),
\end{equation}
where $N$ is the count of Gaussians overlapping the pixel, and $c_i$ is the color from SH coefficients. Gaussians are adaptively adjusted during optimization for accurate scene representation. For details, see~\cite{kerbl3Dgaussians}.

Scaffold-GS~\cite{lu2024scaffold} extends 3D-GS by introducing a structured and sparse hierarchical representation initialized from a COLMAP point cloud $\mathbf{P} \in \mathbb{R}^{M \times 3}$. The points are voxelized into anchors:
\begin{equation}
    \mathbf{V} = \left\{\left\lfloor\frac{\mathbf{P}}{\epsilon}\right\rceil\right\} \cdot \epsilon,
\end{equation}
where $\epsilon$ is the voxel size. Each anchor \( v \in \mathbf{V} \) is characterized by a feature \( f_v \), a scale \( l_v \), and \( k \) learnable offsets \( \mathbf{O}_v \). To achieve multi-resolution and view-dependent modeling, a feature bank \( \{ f_v, f_{v_{\downarrow_1}}, f_{v_{\downarrow_2}} \} \) is constructed, consisting of the original anchor feature and two downsampled versions at different levels. For a given camera position \( \mathbf{x}_c \), the distance \( \delta_{vc} \) and direction \( \vec{\mathbf{d}}_{vc} \) between the anchor and the camera are computed, where \( \delta_{vc} \) represents the Euclidean distance between the anchor and the camera, and \( \vec{\mathbf{d}}_{vc} \) is the normalized displacement vector indicating the viewing direction from the camera to the anchor. These spatial relationships are then processed through a multi-layer perceptron (MLP), which predicts the blending weights \( \{w, w_1, w_2\} \). The weights are normalized using the softmax function and are applied to the feature bank to obtain the final blended feature \( \hat{f_v} \) as follows:
\begin{equation}
    \hat{f_v} = w \cdot f_v + w_1 \cdot f_{v_{\downarrow_1}} + w_2 \cdot f_{v_{\downarrow_2}}
\end{equation}

This structured approach dynamically refines the attributes of Gaussians during training to better capture scene geometry and view-dependent details while reducing redundancy and maintaining high rendering quality.
\subsection{PEP-GS}
\label{sec:3.2}

\subsubsection{View-Adaptive Color Refinement with Attention Mechanisms}

To capture complex view-dependent appearance changes under varying lighting and viewing conditions, we propose Hierarchical Granular-Structural Attention (HGSA), which integrates fine-grained feature enhancement and structural-level attention to dynamically aggregate features. This hierarchical mechanism adapts to directional changes and effectively combines detail-sensitive and global context information to optimize color representation.

Given an input feature \( \mathbf{f}_{in} \in \mathbb{R}^{N \times (32+3+1)} \), which is formed by concatenating the viewing direction $\delta_{vc}$, the relative distance $\vec{\mathbf{d}}_{vc}$, and the integrated anchor feature $\hat{f}_v$:
\begin{align}
   \mathbf{f}_{in} = \delta_{vc} \oplus \vec{\mathbf{d}}_{vc} \oplus \hat{f}_v,
    \label{eq2}
\end{align}
where $\oplus$ denotes the operation of concatenation. As shown in Eq.~\ref{eq2}, we first apply average pooling separately along two directions: horizontally (along the height) and vertically (along the width). These pooled features are then processed through a 1D convolutional layer, followed by a custom normalization and a sigmoid activation function to generate positional attention maps in both directions \( y^h \) and \( y^w \).

\par
\begin{align}
  z^h(h) = \frac{1}{ H}{\sum_{0\leq i<H} \mathbf{f}_{in}{(h,i)}},\quad
  z^w(w) = \frac{1}{ W}{\sum_{0\leq j<W} \mathbf{f}_{in}{(j,w)}}.
    \label{eq2}
\end{align}
where the input feature \( \mathbf{f}_{in} \) is a concatenation of three tensors: the anchor feature \( \hat{f}_v \), the relative viewing distance \( \delta_{vc} \), and the direction \( \vec{\mathbf{d}}_{vc} \) between the camera and the anchor point. These attention maps are then multiplied element-wise with the original input feature tensor \( \mathbf{f}_{\text{in}} \), yielding refined granular regional feature \( \mathbf{f}_{\text{gra}} \). 

\begin{equation}
    \mathbf{f}_{gra} = \mathbf{f}_{in} \times y^h \times y^w,
        \label{eq11}
\end{equation}

Next, we apply structural-level attention to capture global dependencies and encode structural relationships across the input feature space. This attention mechanism uses a multi-head attention framework, where the granular regional feature \( \mathbf{f}_{\text{gra}} \) is first projected into query \( Q \), key \( K \), and value \( V \) spaces using learnable weight matrices. The attention mechanism computes contextual relevance between features by applying the scaled dot-product attention, and the resulting attention weights are used to aggregate the global context into the output feature \( \mathbf{O} \):

\begin{equation}
\mathbf{O} = \operatorname{Softmax}\left(\frac{Q K^T}{\sqrt{d_{\text{head}}}}\right) V,
\end{equation}

The multi-head outputs are concatenated and projected back to the original input dimension. To ensure stability, residual connections and LayerNorm are applied:
\begin{equation}
\mathbf{f}_{\text{HGSA}} = \operatorname{LayerNorm}(\mathbf{O} + \mathbf{f}_{\text{gra}}).
\end{equation}

The structural-level attention mechanism plays a pivotal role in encoding global dependencies, allowing the model to capture long-range interactions and maintain structural consistency across the feature space. By aggregating context-aware information, it complements the granular component, resulting in a unified representation that effectively integrates fine-grained details with global structure. This integration enables robust optimization of view-dependent color refinement in complex 3D scenes, particularly in scenarios with intricate lighting and occlusions.

\subsubsection{Robust Framework for Gaussian Attribute Estimation}

The traditional multilayer perceptron (MLP) is a cornerstone in deep learning, widely used for approximating multivariable functions through a series of nonlinear mappings. An MLP consists of weight matrices \(W\) and activation functions \(\sigma\) applied in alternation, formally expressed as:
\begin{equation}
\text{MLP}(Z) = (W_{K-1} \circ \sigma \circ W_{K-2} \circ \sigma \circ \dots \circ W_1 \circ \sigma \circ W_0) Z,
\label{eq:mlp}
\end{equation}
where \(Z\) is the input, \(W_i\) are weight matrices, and \(\sigma\) denotes activation functions. Despite its versatility, the application of MLPs in high-fidelity 3D reconstruction tasks is hindered by their computational complexity and inability to handle high-dimensional features effectively. This limits their capacity to model intricate geometries, fine-scale textures, and subtle color variations, leading to issues such as shadow artifacts, loss of structural consistency in thin geometries, and suboptimal color rendering in challenging lighting conditions.

In contrast, the Kolmogorov-Arnold Network (KAN), grounded in the Kolmogorov-Arnold representation theorem, introduces a modular and interpretable alternative. Unlike traditional MLPs with fixed activation functions, KAN utilizes learnable activation functions, enhancing the network’s adaptability. A \(K\)-layer KAN network is defined as a nested composition of learnable activation layers:
\begin{equation}
\operatorname{KAN}(\mathbf{Z})=\left(\boldsymbol{\Phi}_{K-1} \circ \boldsymbol{\Phi}_{K-2} \circ \cdots \circ \boldsymbol{\Phi}_{1} \circ \boldsymbol{\Phi}_{0}\right) \mathbf{Z},
\label{eq:kan}
\end{equation}
where each layer \(\boldsymbol{\Phi}_i\) contains \(n_{\text{in}} \times n_{\text{out}}\) learnable activation functions \(\phi\), as shown below:
\begin{equation}
\boldsymbol{\Phi}=\left\{\phi_{k, q, p}\right\}, \quad p=1,2, \cdots, n_{\text{in}}, \quad q=1,2, \cdots, n_{\text{out}}.
\label{eq:phi}
\end{equation}

The pre-activation of \(\phi_{k,q,p}\) is \(Z_{k,p}\), while its post-activation is given by:
\begin{equation}
\tilde{Z}_{k,q,p}=\phi_{k,q,p}(Z_{k,p}).
\label{eq:activation}
\end{equation}

The activation value of the \((k+1,q)\)-th neuron is computed as:
\begin{equation}
Z_{k+1,q} = \sum_{p=1}^{n_k} \tilde{Z}_{k,q,p} = \sum_{p=1}^{n_k}\phi_{k,q,p}(Z_{k,p}), \quad q=1,\cdots,n_{\text{out}}.
\label{eq:neuron}
\end{equation}

Building on this foundation, KAN replaces traditional MLPs for processing input feature \(\mathbf{f}_{in}\). For opacity prediction, a Tanh activation function is applied to the KAN output, with thresholds set to retain essential neural Gaussian distributions ( $\alpha \geq \tau_\alpha$) while preventing premature removal of critical components. KAN dynamically adjusts the scale of each neural Gaussian distribution based on anchor features, viewing angle, and position, preserving fine-scale textures and thin geometries. Similarly, KAN predicts rotation parameters, allowing neural Gaussian distributions to adapt their orientation based on the viewing angle, enhancing the capture of spatial relationships and view-dependent occlusions.

KAN’s ability to precisely control opacity, scaling, and rotation significantly improves reconstruction quality, ensuring better preservation of intricate textures, thin structures, and fine details. By addressing the limitations of traditional MLPs and Gaussian-based methods, KAN enhances stability and adaptability in view-dependent scenarios, particularly in handling complex lighting, weak textures, and dynamic occlusions.

\subsubsection{Loss Fuction}

Our overall loss function is designed as:
\begin{equation}
\begin{split}
\mathcal{L} = & \big((1 - \lambda_{\text{D-SSIM}}) \cdot \mathcal{L}_2  
+ \lambda_{\text{D-SSIM}} \cdot \mathcal{L}_{\text{D-SSIM}} 
+ \lambda_{\text{vol}} \cdot \mathcal{L}_{\text{vol}}\big) \\
& \cdot (1 - \lambda_{\text{NLPD}}) 
+ \lambda_{\text{NLPD}} \cdot \mathcal{L}_{\text{NLPD}},
\end{split}
\end{equation}

To regulate the volume of neural Gaussians, the regularization term is defined as \( \mathcal{L}_{\text{vol}} = \sum_{i=1}^{N_{\text{ng}}} \operatorname{Prod}(s_i) \), where \( N_{\text{ng}} \) indicates the total count of neural Gaussians in the scene, and \( \operatorname{Prod}(\cdot) \) calculates the product of the components of a vector, specifically the scale \( s_i \) for each Gaussian in this scenario. In particular, \( \mathcal{L}_{\text{NLPD}} \) plays a critical role in maintaining multi-scale perceptual consistency.

The Normalized Laplacian Pyramid Distance (NLPD) loss \( \mathcal{L}_{\text{NLPD}} \) is designed to measure multi-scale perceptual differences between the rendered image and the ground truth by leveraging a Laplacian pyramid decomposition to create a multi-resolution representation. At each scale \( i \), the residual \( R_i \) is computed as the difference between the original image and its reconstructed version. These residuals are then normalized using divisive normalization:

\begin{equation}
    R_i^{\text{norm}} = \frac{R_i}{\sigma_i + F(R_i)}
\end{equation}
where \( \sigma_i \) is a scale-dependent noise parameter, and \( F(\cdot) \) is a local filter applied to the residual magnitudes, ensuring robustness to variations in texture and noise.

The final perceptual difference is obtained by aggregating the squared differences of the normalized residuals between the rendered image and the ground truth across all scales:
\begin{equation}
    \mathcal{L}_{\text{NLPD}} = \frac{1}{k} \sum_{i=1}^{k} \sqrt{ \frac{1}{N_i} \sum_{j=1}^{N_i} \left( R_i^{x, \text{norm}}[j] - R_i^{y, \text{norm}}[j] \right)^2 }
\end{equation}
where \( N_i \) represents the number of pixels at scale \( i \), and \( k \) is the total number of scales considered.

This perceptual loss effectively captures fine-grained details and texture differences across multiple resolutions, enhancing both global structural similarity and local detail consistency. When integrated with other loss components, \( \mathcal{L}_{\text{NLPD}} \) ensures perceptually consistent renderings that closely align with the ground truth, even in challenging scenarios involving intricate textures and lighting effects.

\section{Experiments}
\label{sec:exp}
\subsection{Setups}
\subsubsection{\textbf{Datasets and Metrics}}
We conduct a comprehensive evaluation of our proposed model on seven scenes from Mip-NeRF360~\cite{barron2022mipnerf360}, two scenes from Tanks\&Temples~\cite{Knapitsch2017}, two scenes from DeepBlending~\cite{hedman2018deep}. Consistent with methodologies outlined in previous works~\cite{barron2022mipnerf360,kerbl3Dgaussians}, we adopted a train/test split approach where every 8th photo was selected for testing. To assess the rendering quality, we further measured the average Peak Signal-to-Noise Ratio (PSNR), Structural Similarity Index Measure (SSIM)~\cite{1284395}, and Learned Perceptual Image Patch Similarity (LPIPS)~\cite{zhang2018unreasonable}. 

\subsubsection{\textbf{Implementation Details}}
To maintain fairness in evaluation, we configure \(k=10\) for all experiments, consistent with the settings used in Scaffold-GS. In the design of the loss function, we assign a value of 0.2 to \(\lambda_{\text{NLPD}}\). For the HGSA, the multi-head attention mechanism is configured with 7 heads. We assess the quality of our approach by comparing it to current state-of-the-art(SOTA) baselines, including Instant-NGP~\cite{mueller2022instant}, Plenoxels~\cite{fridovich2022plenoxels}, Mip-NeRF360~\cite{barron2022mipnerf360}, 3D-GS, Mip-splatting~\cite{yu2023mip} and Scaffold-GS. In line with the approach described in 3D-GS, our models undergo training for 30K iterations across all scenes.

\subsection{Comparisons}
\subsubsection{\textbf{Quantitative Comparisons.}} 
\begin{figure}[!ht]
    \centering
    \includegraphics[width=1\linewidth]{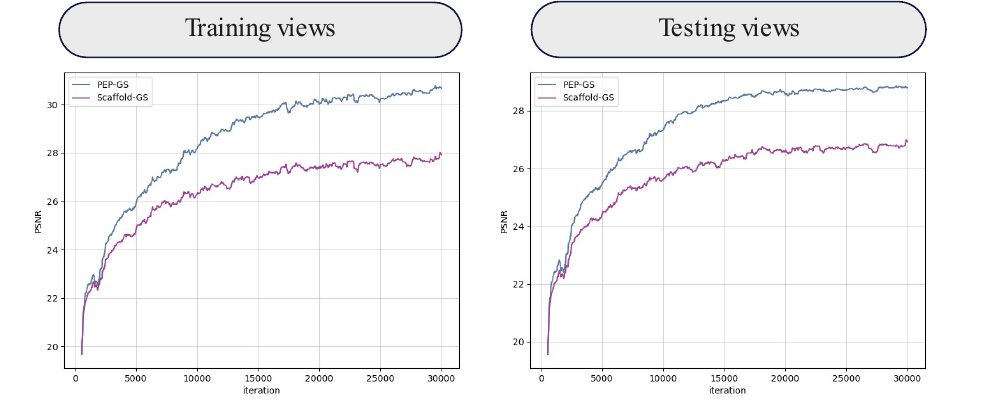}
    \vspace{-7mm}
    \caption{The PSNR curve of Scaffold-GS and PEP-GS across the truck scene in the Tanks \& Temples dataset under training and test views.}
    \label{fig:graph}
\end{figure}

Table.~\ref{tab:real-world datasets} presents the results on three real-world datasets~\cite{barron2022mipnerf360,Knapitsch2017,hedman2018deep}, highlighting that our approach achieves competitive performance in rendering quality compared to state-of-the-art methods and stands out in terms of image quality and perceived similarity. While Scaffold-GS achieves the highest PSNR performance on the Deep Blending dataset, our method demonstrates more consistent and robust performance across other datasets, with PEP-GS notably outperforming Scaffold-GS on the Tanks\&Temples dataset, as illustrated in Figure.~\ref{fig:graph}. It can be noticed that our approach achieves comparable results with 3D-GS on Mip-NeRF360 dataset, and achieves superior results in Tanks\&Templates datasets, exhibiting notable competitiveness in terms of SSIM and LPIPS.


\begin{table*}[!ht]
        \centering
        \caption{Quantitative comparison to previous methods on real-world datasets. Competing metrics are extracted from respective papers.}

        \label{tab:real-world datasets}
    \begin{tabular}{C{2.8cm}|C{1.2cm}C{1.2cm}C{1.2cm}|C{1.2cm}C{1.2cm}C{1.2cm}|C{1.2cm}C{1.2cm}C{1.2cm}} 
\hline\hline
Dataset     & \multicolumn{3}{c|}{Mipnerf-360} & \multicolumn{3}{c|}{Tanks\&Templates} & \multicolumn{3}{c}{Deep Blending} \\ 
\begin{tabular}{c|c} Method & Metrics \end{tabular}      & PSNR$\uparrow$   & SSIM$\uparrow$  & LPIPS$\downarrow$   & PSNR$\uparrow$   & SSIM$\uparrow$  & LPIPS$\downarrow$  & PSNR$\uparrow$   & SSIM$\uparrow$  & LPIPS$\downarrow$  \\ \hline
Instant-NGP   &26.43 &0.725 &0.339 
              &21.72 &0.723 &0.330 
              &23.62 &0.797 &0.423  \\
Plenoxels     &23.62 &0.670 &0.443 
              &21.08 &0.719 &0.379 
              &23.06 &0.795 &0.510  \\
Mip-NeRF360   &29.23 &0.844 &0.207 
              &22.22 &0.759 &0.257 
              &29.40 &0.901 &\cellcolor{2}0.245  \\           
3D-GS         &28.69 &\cellcolor{1}0.870 &\cellcolor{1}0.182 
              &23.14 &0.841 &0.183 
              &29.41 &\cellcolor{3}0.903 &\cellcolor{1}0.243  \\
Mip-Splatting &\cellcolor{3}29.12 &\cellcolor{2}0.869 &\cellcolor{3}0.184 
              &\cellcolor{3}23.79 &\cellcolor{3}0.848 &\cellcolor{3}0.176 
              &\cellcolor{3}29.57 &0.900 &\cellcolor{2}0.245  \\   
Scaffold-GS   &\cellcolor{2}29.35 &\cellcolor{1}0.870 &0.188 
              &\cellcolor{2}24.14 &\cellcolor{2}0.853 &\cellcolor{2}0.175
              &\cellcolor{1}30.26 &\cellcolor{2}0.909 &0.252  \\ \hline
PEP-GS (Ours) &\cellcolor{1}29.80 &\cellcolor{1}0.870 &\cellcolor{2}0.183 
              &\cellcolor{1}24.79 &\cellcolor{1}0.862 &\cellcolor{1}0.163 
              &\cellcolor{2}30.00 &\cellcolor{1}0.910 &\cellcolor{3}0.249  \\

\hline\hline
\end{tabular}
\end{table*}

\begin{figure*}[t!]
	\centering
	\includegraphics[width=\linewidth]{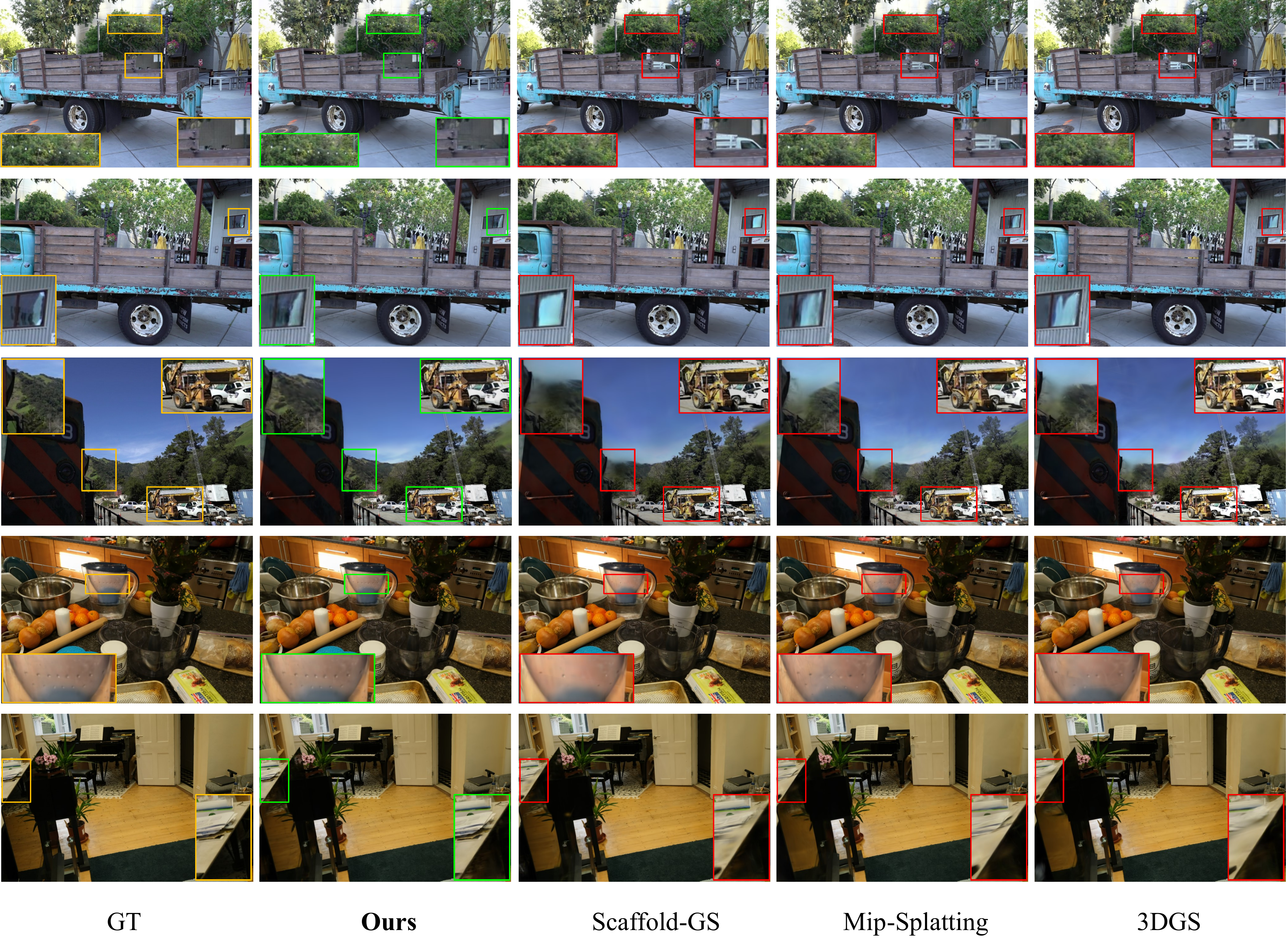}
	\vspace{-20pt}
	\caption{Compared to existing baselines, our method demonstrates superior detail preservation, reduced artifacts, and improved color consistency. Notably, in the second row of the images, our method also shows certain improvements in handling specular reflections compared to other approaches.}
	\vspace{-10pt}
	\label{fig:qualitative}
\end{figure*}

\subsubsection{\textbf{Qualitative Comparisons.}}
\begin{figure*}[t]
    \centering
    \includegraphics[width=1\linewidth]{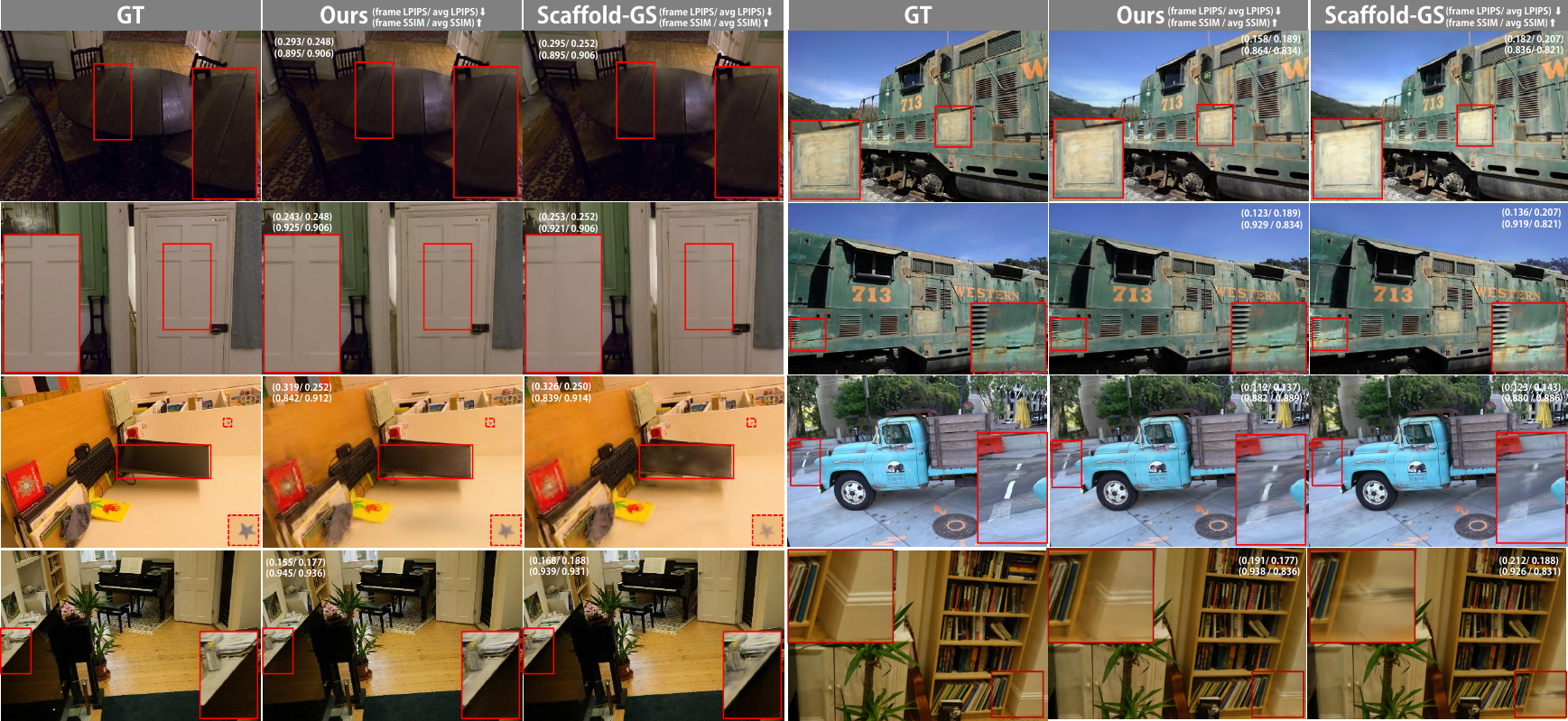}
    \caption{A systematic qualitative comparison between Scaffold-GS and PEP-GS is carried out across the Mip-NeRF360, Tanks\&Temples, and Deep Blending datasets to highlight their relative strengths in diverse real-world scenarios.}
    \label{fig:comp2sca}
\end{figure*}

Qualitative results across various datasets are provided in Figure.~\ref{fig:qualitative}. 
We also conducted a more comprehensive and detailed comparison of rendering results against the scaffold method. As shown in the first and second rows of the left column in Figure.~\ref{fig:comp2sca}, it is evident that our method demonstrates sharper details and textures, effectively preserving local structures and fine-scale details during image reconstruction. As shown in the first and second rows of the right column in Figure.~\ref{fig:comp2sca}, under challenging sunlight environments, the Scaffold-GS method exhibits slightly excessive brightness, leading to the loss of certain details. In contrast, our method not only accurately maintains clearer textures and structural features but also avoids overexposure or shadow detail loss. Figure.~\ref{fig:Intro} further illustrate that in challenging scenes like grass fields, our method better suppresses artifacts and unnatural textures while adeptly capturing complex textures and thin structures, exhibiting greater robustness. In terms of metric analysis, our method achieves lower LPIPS values and higher SSIM scores, confirming its significant advantage in overall visual quality and perceptual consistency.

\begin{figure*}[t!]
	\centering
	\includegraphics[width=\linewidth]{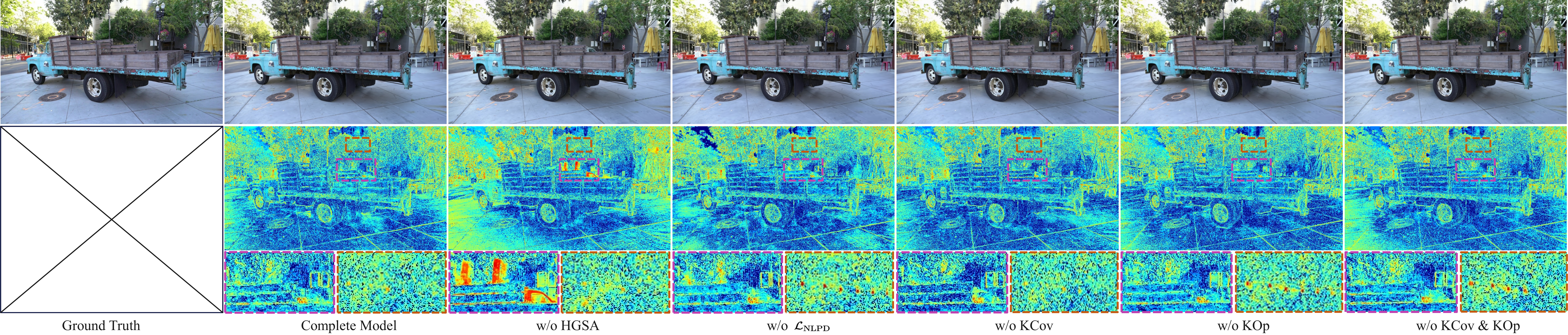}
	\vspace{-20pt}
	\caption{Ablation studies. The top images show our rendering results, and the bottom images present the error maps computed between these renderings and the ground truth (GT). In these maps, more intense colors denote larger discrepancies from the GT. Similar to Table.~\ref{tab:Mipnerf360 datasets}, KOp and KCov represent two methods that utilize the KAN framework to predict Gaussian opacity and covariance attributes, respectively.}
	\vspace{-10pt}
	\label{fig:ablation}
\end{figure*}

\begin{table}[htbp]
        \centering
        \caption{An ablation study evaluates the impact of removing components from PEP-GS.} 
        \label{tab:Mipnerf360 datasets}
    \begin{tabular}{p{3cm}|C{1.2cm}C{1.2cm}C{1.2cm}}
\hline\hline

Model setting      & PSNR$\uparrow$   & SSIM$\uparrow$  & LPIPS$\downarrow$ \\ \hline
1) w/o $\mathcal{L}_{\text{NLPD}}$     &32.60 &0.935 &0.184   \\
2) w/o HGSA                           &32.14 &0.934 &0.179   \\
3) w/o KCov                            &33.10 &0.936 &0.178   \\
4) w/o KOp                            &33.13 &0.936 &0.179   \\
5) w/o both 3) and 4)                  &32.99 &0.935 &0.180   \\ \hline
Our Complete Model        & \textbf{33.19} & \textbf{0.936} & \textbf{0.177} \\
\hline\hline
\end{tabular}
\end{table}

\subsection{Ablation Study}
To assess the effectiveness of each component of PEP-GS, we conducted ablation studies on the \emph{room} scene from the Mip-NeRF360 dataset and the \emph{truck} scene from the Tanks\&Temples dataset. The quantitative and qualitative results are reported in Table.~\ref{tab:Mipnerf360 datasets} and Figure.~\ref{fig:ablation}, respectively. We summarize our findings here. 

\textbf{HGSA}. The removal of HGSA results in a marked degradation in both quantitative metrics and perceptual quality. As shown in Figure.~\ref{fig:ablation}, the rendered images result in false geometry without HGSA, introducing visual artifacts that were not present in the original scene. This degradation underscores the importance of HGSA in effectively capturing local geometric features and view-dependent effects. Its absence highlights the challenges of modeling intricate view-dependent phenomena in 3D scenes. 

\textbf{KCov\&KOp}. The exclusion of either KCov or KOp alone leads to marginal decreases in performance metrics. However, removing both components results in a significant reduction in rendering quality. More importantly, the rendered images lack critical fine-grained details and exhibit poor representation of thin and delicate structures, such as small light bulbs or intricate textures. KCov and KOp collectively enhance the model’s ability to predict Gaussian attributes dynamically and precisely. These components ensure adaptive scaling and spatially consistent rendering, enabling the system to faithfully represent subtle geometric features and avoid over-simplification of the scene. 

\textbf{$\mathcal{L}_{\text{NLPD}}$}. Excluding $\mathcal{L}_{\text{NLPD}}$ results in perceptual inconsistencies similar to those observed when KCov and KOp are removed. The absence of this multi-scale perceptual loss disrupts the model’s ability to align visual features across different spatial resolutions, leading to the loss of fine-scale details and textures, such as thin structures and small objects. $\mathcal{L}_{\text{NLPD}}$ plays a critical role in ensuring the fidelity of reconstructed textures and preserving structural details, particularly in scenes with complex geometric and photometric properties.




\section{Conclusion and Limitation}
\label{sec:conclusion}
In this paper, we propose PEP-GS, a perceptually-enhanced framework for structured 3D Gaussian splatting, addressing key challenges in view-adaptive rendering. Our method introduces a Hierarchical Granular-Structural Attention mechanism to better capture complex view-dependent effects, and utilizes Kolmogorov-Arnold Networks for robust and interpretable Gaussian attribute estimation. Additionally, the incorporation of the Normalized Laplacian Pyramid Distance loss enhances perceptual consistency across views, ensuring high-quality renderings with superior texture fidelity and geometric precision. 

Even though experimental results show that PEP-GS surpasses existing methods, excelling in handling complex lighting and fine-scale details, there are still some limitations. Compared with scaffold-GS, while our method achieves superior metric scores and rendering performance, it slightly compromises rendering efficiency. We anticipate that our algorithm will progressively improve its rendering speed as the field advances.

\section{Acknowledgments}
This work is supported by the Hainan Provincial Natural Science Foundation of China (Grant No. 722RC678).

\appendix





\newpage
\bibliographystyle{named}
\bibliography{ijcai25}

\newpage
\appendix
\section*{Appendix}

\begin{figure*}[t!]
	\centering
	\includegraphics[width=0.75\linewidth]{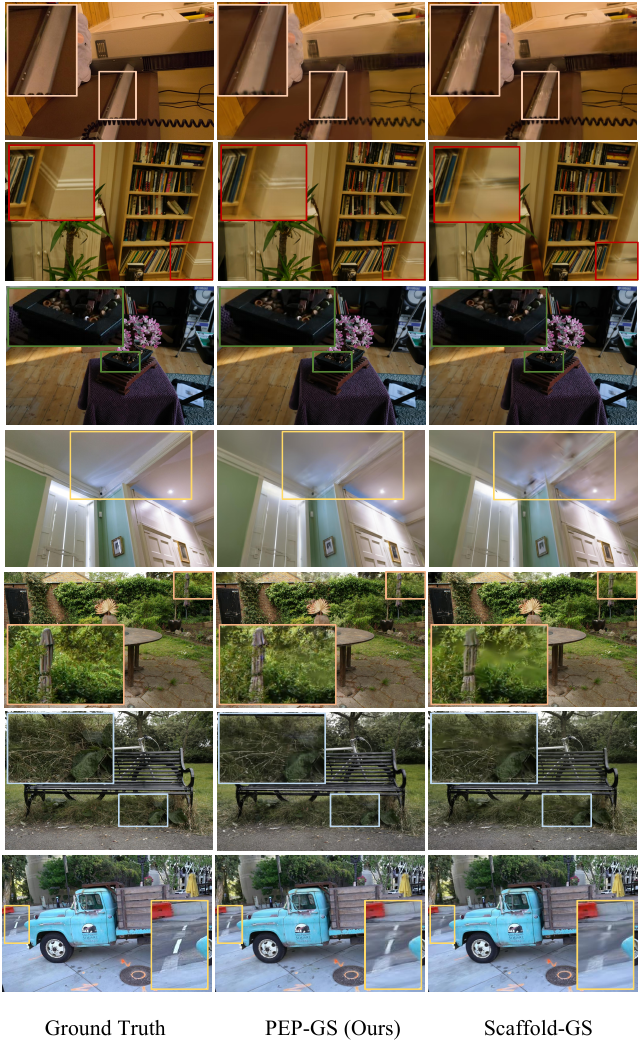}
	\vspace{-5pt}
	\caption{Qualitative comparison between PEP-GS (Ours) and Scaffold-GS.}
	\vspace{-5pt}
	\label{fig:appendix}
\end{figure*}

We present comprehensive evaluation metrics for all scenes from the three public datasets mentioned in the main text. As shown in Tables.~\ref{table:appendix_tan_dt} and \ref{table:appendix_mip360}, our model consistently achieves superior performance across the majority of scenes. These quantitative results align with the enhanced perceptual capabilities of our model, further illustrated in Figures.~\ref{fig:Comparison of texture-less regions and thin structures} and \ref{fig:appendix}.

In indoor scenes such as Playroom and Dr Johnson, our method excels in handling complex lighting conditions and preserving fine geometric details, areas where traditional methods often struggle. The integration of our Hierarchical Granular-Structural Attention (HGSA) mechanism plays a crucial role in maintaining consistency across different viewing angles while effectively preserving intricate texture details.

Similarly, in outdoor environments like Truck and Garden, our approach demonstrates a remarkable capability in managing natural lighting variations and capturing complex geometric structures with high fidelity. The effectiveness of our Kolmogorov-Arnold Networks (KAN) for opacity and covariance estimation is particularly evident in challenging scenarios involving thin structures and texture-less regions. This component ensures precise control over Gaussian attributes, leading to more accurate geometric reconstruction and improved preservation of fine-scale details.

\begin{table}[!htbp]
    \centering
    \small
    \caption{Quantitative performance comparison across Tanks\&Temples and Deep Blending scenes.}
    \label{table:appendix_tan_dt}
    \resizebox{\columnwidth}{!}{
    \begin{tabular}{c|cc|cc}
    \hline\hline

   \multicolumn{5}{c}{PSNR$\uparrow$} \\
    \hline
    \begin{tabular}{c|c} Method & Scenes \end{tabular}      & Truck & Train & Dr Johnson & Playroom \\
    \hline
    Instant-NGP         & 23.26 & 20.17 & 27.75 & 19.48 \\
    Plenoxels           & 23.22 & 18.93 & 23.14 & 22.98 \\
    Mip-NeRF360         & 24.91 & 19.52 & 29.14 & 29.66 \\
    3D-GS               & 25.19 & 21.10 & 28.77 & 30.04 \\
    Mip-Splatting       & 25.56 & 22.02 & 29.11 & 30.03 \\
    Scaffold-GS         & 25.82 & 22.46 & \textbf{29.68} & \textbf{30.84} \\
    \hline
   PEP-GS (Ours)        & \textbf{26.34} & \textbf{23.24} & 29.33 & 30.66 \\
    \hline
     \multicolumn{5}{c}{SSIM$\uparrow$} \\
    \hline
    \begin{tabular}{c|c} Method & Scenes \end{tabular}      & Truck & Train & Dr Johnson & Playroom \\
   \hline
    Instant-NGP              & 0.779 & 0.666 & 0.899 & 0.906 \\
    Plenoxels                & 0.774 & 0.663 & 0.787 & 0.802 \\
    Mip-NeRF360              & 0.857 & 0.660 & 0.901 & 0.900 \\
    3D-GS                    & 0.879 & 0.802 & 0.899 & 0.906 \\
    Mip-Splatting            & 0.881 & 0.816 & 0.899 & 0.902 \\
    Scaffold-GS              & 0.886 & 0.821 & \textbf{0.906} & 0.912 \\
    \hline
   PEP-GS (Ours)             & \textbf{0.889} & \textbf{0.834} & \textbf{0.906} & \textbf{0.914} \\
    \hline
    \multicolumn{5}{c}{LPIPS$\downarrow$} \\
    \hline
    \begin{tabular}{c|c} Method & Scenes \end{tabular}      & Truck & Train & Dr Johnson & Playroom \\
    \hline
    Instant-NGP              & 0.274 & 0.386 & 0.381 & 0.465 \\
    Plenoxels                & 0.335 & 0.422 & 0.521 & 0.499 \\
    Mip-NeRF360              & 0.159 & 0.354 & \textbf{0.237} & 0.252 \\
    3D-GS                    & 0.148 & 0.218 & 0.244 & \textbf{0.241} \\
    Mip-Splatting            & 0.147 & 0.205 & 0.245 & 0.245 \\
    Scaffold-GS              & 0.143 & 0.207 & 0.252 & 0.252 \\
    \hline
   PEP-GS (Ours)             & \textbf{0.137} & \textbf{0.189} & 0.248 & 0.250 \\
   \hline\hline
    \end{tabular}
    }
\end{table}

\begin{table*}[!htbp]
    \centering
    \small
    \caption{Quantitative performance comparison across Mip-NeRF 360 scenes.}
    \label{table:appendix_mip360}
    \resizebox{\textwidth}{!}{
        \begin{tabular}{l|ccccccc}
            \hline\hline
            \multicolumn{8}{c}{PSNR$\uparrow$} \\
            \hline
            \begin{tabular}{c|c} Method & Scenes \end{tabular}                        & Bicycle & Garden & Stump & Room & Counter & Kitchen & Bonsai \\
            \hline
            Instant-NGP             & 22.19  & 24.60  & 23.63  & 29.27  & 26.44  & 28.55  & 30.34 \\
            Plenoxels               & 21.91  & 23.49  & 20.66  & 27.59  & 23.62  & 23.42  & 24.67\\
            Mip-NeRF360             & 24.37  & 26.98  & 26.40  & 31.63  & 29.55  & \textbf{32.23}  & 33.46 \\
            3D-GS          & \textbf{25.25}  & 27.41  & 26.55  & 30.63  & 28.70  & 30.32  & 31.98  \\
            Mip-Splatting           & 25.13  & 27.35  & 26.63  & 31.75  & 29.12  & 31.51  & 32.36 \\
            Scaffold-GS             & 25.20  & 27.28  & 26.63  & 32.15  & 29.66  & 31.77  & 32.78 \\
            \hline
            PEP-GS (Ours)          & 25.11  & \textbf{27.61} & \textbf{26.66} & \textbf{33.19} & \textbf{30.26} & 31.87 & \textbf{33.93} \\
            \hline 
            
            \multicolumn{8}{c}{SSIM$\uparrow$} \\
            \hline
            \begin{tabular}{c|c} Method & Scenes \end{tabular}                         & Bicycle & Garden & Stump & Room & Counter & Kitchen & Bonsai \\
            \hline
            Instant-NGP              & 0.491  & 0.649  & 0.574  & 0.855  & 0.798   & 0.818   & 0.890  \\
            Plenoxels                & 0.496  & 0.606  & 0.523  & 0.842  & 0.759  & 0.648  & 0.814  \\
            Mip-NeRF360              & 0.685  & 0.813  & 0.744  & 0.913  & 0.894  & 0.920  & 0.941  \\ 
            3D-GS                    & \textbf{0.771}  & \textbf{0.868}  & \textbf{0.775}  & 0.914  & 0.905  & 0.922  & 0.938  \\ 
            Mip-Splatting            & 0.747  & 0.853  & 0.769  & 0.925  & 0.913  & 0.931  & 0.945  \\
            Scaffold-GS              & 0.746  & 0.850  & 0.765  & 0.931  & \textbf{0.919}  & 0.932  & 0.949  \\
            \hline
            PEP-GS (Ours)            & 0.735  & 0.852  & 0.758  & \textbf{0.936}  & \textbf{0.919}  & \textbf{0.935}  & \textbf{0.955}  \\
            \hline 
            \multicolumn{8}{c}{LPIPS$\downarrow$} \\ 
           
            \hline
            \begin{tabular}{c|c} Method & Scenes \end{tabular}                         & Bicycle & Garden & Stump & Room & Counter & Kitchen & Bonsai \\
            \hline
            Instant-NGP              & 0.487 & 0.312  & 0.450  & 0.301 & 0.342  & 0.254  & 0.227  \\
            Plenoxels                & 0.506  & 0.386  & 0.503  & 0.419  & 0.441  & 0.447  & 0.398  \\
            Mip-NeRF360              & 0.301  & 0.170  & 0.261  & 0.211  & 0.204  & 0.127  & 0.176  \\
            3D-GS                    & \textbf{0.205}  & \textbf{0.103}  & \textbf{0.210}  & 0.220  & 0.204  & 0.129  & 0.205  \\
            Mip-Splatting            & 0.245  & 0.123  & 0.242  & 0.197  & 0.183  & 0.116  & 0.181  \\
            Scaffold-GS              & 0.256  & 0.135  & 0.259  & 0.188  & 0.181  & 0.118  & 0.178  \\
            \hline
            PEP-GS (Ours)            & 0.260  & 0.130  & 0.257  &\textbf{ 0.177}  &\textbf{ 0.176}  &\textbf{ 0.113}  &\textbf{ 0.168}  \\
            \hline\hline
        \end{tabular}
    }
\end{table*}

\end{document}